\documentclass[10pt,twocolumn,letterpaper]{article}

\usepackage[pagenumbers]{cvpr} 

\usepackage{todonotes}

\definecolor{cvprblue}{rgb}{0.21,0.49,0.74}
\usepackage[pagebackref,breaklinks,colorlinks,allcolors=cvprblue]{hyperref}
\usepackage{amsfonts}
\usepackage{mathrsfs}
\usepackage{tabularray}
\usepackage{booktabs}
\usepackage{multirow}
\usepackage{amsmath}
\usepackage{amssymb}
\usepackage{latexsym}
\usepackage{mathtools}
\usepackage{xcolor}
\usepackage{colortbl}

\def\paperID{1264} 
\def\confName{CVPR}
\def\confYear{2025}

\title{HeatFormer: A Neural Optimizer for Multiview Human Mesh Recovery}

\author{Yuto Matsubara \qquad
Ko Nishino \\
Graduate School of Informatics, Kyoto University\\
\texttt{\small \href{https://vision.ist.i.kyoto-u.ac.jp/research/heatformer/}{https://vision.ist.i.kyoto-u.ac.jp/research/heatformer/}}
}

\begin{document}
\twocolumn[{
  \maketitle 
  \begin{center}
    \captionsetup{type=figure}

    \includegraphics[width=\linewidth]{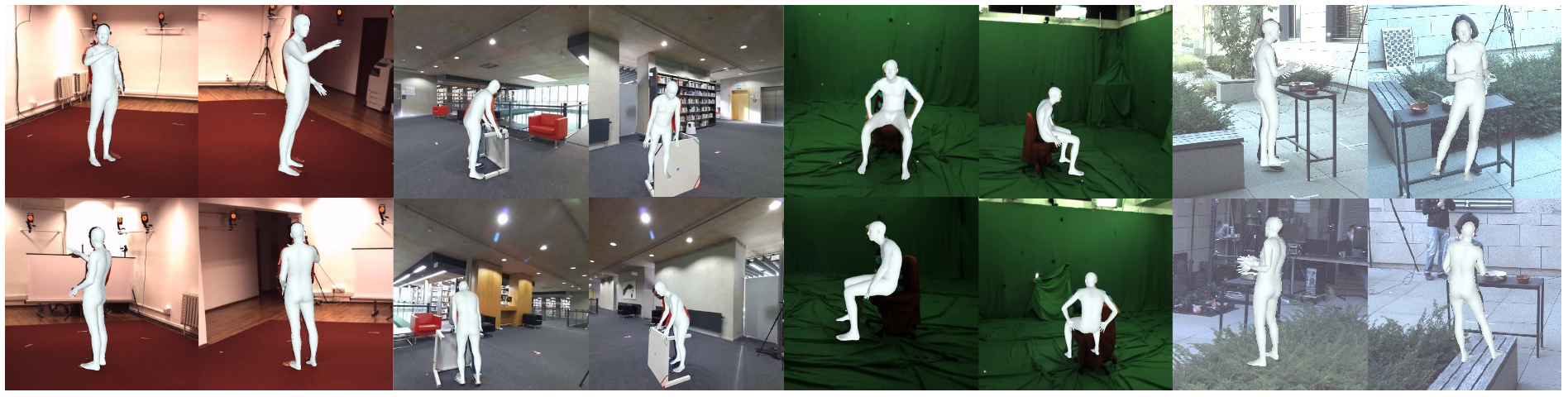}
    
    \captionof{figure}{
    HeatFormer is a novel neural optimizer for human mesh recovery from static multiview images. It recovers the SMPL shape and pose parameters by fully leveraging the views to resolve occlusions (\eg, note how well the body parts occluded by objects and other body parts are recovered). It is also agnostic to the configuration and number of views which is essential for generalization in real-world scenes. (From left to right, the datasets are  Human3.6M \cite{6682899}, BEHAVE \cite{bhatnagar22behave}, MPI-INF-3DHP \cite{mono-3dhp2017}, and RICH \cite{Huang:CVPR:2022}.)}
    \label{fig:opening fi}
\end{center}
}]

\begin{abstract}
We introduce a novel method for human shape and pose recovery that can fully leverage multiple static views. We target fixed-multiview people monitoring, including elderly care and safety monitoring, in which cameras can be installed at the corners of a room or an open space but whose configuration may vary depending on the environment. 
Our key idea is to formulate it as neural optimization. We achieve this with HeatFormer, a neural optimizer that iteratively refines the SMPL parameters given multiview images. HeatFormer realizes this SMPL parameter estimation as heatmap generation and alignment with a novel transformer encoder and decoder. Our formulation makes HeatFormer fundamentally agnostic to the number of cameras, their configuration, and calibration. We demonstrate the effectiveness of HeatFormer including its accuracy, robustness to occlusion, and generalizability through an extensive set of experiments. We believe HeatFormer can serve a key role in passive human behavior modeling. 

\end{abstract}

\section{Introduction}
\label{sec:intro}

Many computer vision applications rely on accurate estimates of the human body in images and videos. 
Strong statistical priors on the body shape (\eg, SMPL \cite{Loper:SMPL}) have enabled significant reduction in the search space and, as a result, robust body estimation from single images. 
By also combining priors on the body pose and camera motion estimation from static background, recent methods have extended human body and pose estimation to dynamic monocular cameras. 
These methods open new possibilities for VR/AR applications, allowing one to place and interact with people observed in the real-world in their own virtual world.

Accurate human body shape and pose recovery is fundamental for static camera applications, too. For instance, a room equipped with a handful of cameras installed in the corners can let us learn about the activities in the room and their people. Automation of this process can lay the foundation for ensuring the well-being of them for security and cognitive monitoring, and enable robotic intervention with co-robots and service robots. A representative need for this is elderly care, in which 24/7 passive monitoring to decide when and how to intervene would be essential. 
Our focus is on this fixed, multiview monitoring which finds many applications in a wide range of disciplines including security, commerce, communication, and entertainment.

One would be tempted to think that the problem is easier in such scenarios as the cameras do not move and we have multiple of them. Why not just apply the successful monocular methods and be done with it? The world, however, is messy. Even in a simple living room, a person can be occluded by all sorts of things, the chair, the table, the sofa, and more often than that the person herself. Internet monocular (dynamic) views targeted in previous approaches usually have unoccluded views, because that is the exact intent of the person capturing the images and videos. In contrast, in daily rooms and street corners, each view would have frequent and severe occlusions as people are their to interact with the environment with their bodies.

From a monocular view, there is no way to know how the occluded body part is shaped and posed. Past methods have leveraged strong learned priors, such as a diffusion model \cite{Stathopoulos_2024_CVPR}, to generate and fill-in these unknowns. This may succeed, but this Bayesian reliance on prior knowledge only hallucinates the unseen with what it think it sees. 
Occluded body parts in one camera may be visible in others but from different vantage points. Can't we leverage them without solely relying on pre-learned priors which inevitably favor ``typical'' configurations? Can't they be fully leveraged by referencing them in the very process of the body shape and pose recovery, not just once as input as in past feed-forward network-based methods?

One needs to be careful not to rely on knowledge of the camera configuration. We do not want to assume that the extrinsic coordinates and directions of the views are prefixed, as that would significantly limit the applications. We want to be able to just install some number of cameras in a room and still fully leverage them to accurately estimate the shape and pose of people that enter the room. 

How do we approach this multiview human shape and pose recovery that is occlusion-robust, multiview-aware, and view configuration flexible? Our key idea is to formulate it as neural optimization. By deriving a novel neural network that learns to iteratively revise the SMPL parameters of a person captured from multiple views, rather than learning to output the parameters in one inference pass, we arrive at a multiview human body shape and pose recovery method that is agnostic to the scene, number of cameras, and their configuration yet that can make the best out of the complementary observations the multiview setting offers.

Our neural human body shape and pose optimizer, which we refer to as HeatFormer, consists of three key technical contributions. The first is the overall architecture. We build upon the Transformer architecture and derive a neural optimizer that iteratively refines the SMPL parameters given multiview images which is fundamentally agnostic to the number of views and their configurations. The second is the formulation of body shape and pose estimation as heatmap matching. 
Directly evaluating the error of SMPL parameters in the image space is near impossible as we do not have masks or accurate texture of the person. Keypoint supervision of joints is more realistic, but are hard to estimate with a neural optimizer as they do not offer dense spatial gradients. We turn to the original idea of heatmaps and recast SMPL parameter estimation as heatmap generation and matching. Finally, we derive a novel Transformer encoder that takes in the joint heatmaps and consolidates their spatial coordination into a single heatmap. Our formulation enables application of HeatFormer to uncalibrated cameras. At the same time, when the cameras are calibrated, HeatFormer fully leverages the camera extrinsics.

We validate the effectiveness of HeatFormer through extensive experiments. We evaluate the accuracy of HeatFormer by comparing it with the current state-of-the-art methods for both monocular and multiview human mesh recovery on major multiview datasets including Human3.6M \cite{6682899} and MPI-INF-3DHP \cite{mono-3dhp2017}.
We quantitatively evaluate the robustness of our method to occlusion on the BEHAVE dataset \cite{bhatnagar22behave}.
Through ablation studies and cross-dataset validation, we evaluate the effectiveness of our formulation of human mesh recovery as a neural optimization. 
The experimental results demonstrate that HeatFormer is a general and robust multiview human body and shape estimator. We believe HeatFormer opens new avenues of research and applications for human behavior understanding. 

\section{Related Work}
\label{sec:related_work}

Estimation of the shape and pose of a person in images can be formulated as the reconstruction of a water-tight mesh model of the person which is often called human mesh recovery (HMR). We review representative methods for HMR from monocular and multiview images and videos. 

\begin{table}
  \centering
  \scalebox{0.65}{
  \begin{tabular}{@{}lccccc@{}}
    \toprule
    Method & Multiview & Heatmap & Feedback & Occl.-Robust & Generalization\\
    \midrule
    HMR2.0 \cite{goel2023humans}&  &  &  & \checkmark & \checkmark \\
    \midrule
    LVS \cite{shin2020multiviewhumanposeshape}& \checkmark &  &  & \checkmark & \checkmark \\
    PaFF \cite{jia2023paff}& \checkmark &  & \checkmark & \checkmark & \checkmark \\
    U-HMR \cite{li2024humanmeshrecoveryarbitrary}& \checkmark &  &  &  \\
    \midrule
    HeatFormer (Ours) & \checkmark & \checkmark & \checkmark & \checkmark & \checkmark\\
    \bottomrule
  \end{tabular}}
  \caption{Our method (``Ours'') fully leverages multiview images in the estimation process itself (``Feedback'') to achieve ``Occlusion-Robust''ness and ``Generalization'' to various environments and view configurations. It achieves this by aligning the ``Heatmap" of the view-dependent SMPL estimates with those of the input images. We compare with other multiview methods and demonstrate the effectiveness due to these key strengths.}
  \label{tab:example}
\end{table}

\vspace{-8pt}
\paragraph{Single-View Human Mesh Recovery}
Many impressive results have been achieved for HMR from single images \cite{kanazawaHMR18, Pavlakos_2018_CVPR, Moon_2020_ECCV_I2L-MeshNet, goel2023humans}. 
Although some works directly estimate mesh vertices \cite{you2023co, Choi_2020_ECCV_Pose2Mesh, lin2021-mesh-graphormer, lin2021end-to-end, cho2022FastMETRO, kolotouros2019cmr, lekakolyris2024meshpose, fiche2024vq}, most methods estimate the parameters of a learned statistical shape and pose model of the human body, namely SMPL \cite{Loper:SMPL} as it greatly reduces the number of unknowns for this fundamentally ill-posed problem. A network can be learned to regress these SMPL parameters from an image. Kanazawa et al. \cite{kanazawaHMR18} introduced a discriminative prior on the body pose to further tame the solution space for such feed-forward regression. 

HMR can also be formulated as optimization, \ie, gradual fitting of the mesh model by iterative refinement. This enables the method to reference the visual evidence, often in the form of 2D joints, to adapt to the body shape and pose instantiation. SPIN \cite{kolotouros2019spin} combines feed-forward regression (a ConvNet) for initial estimates and optimization (SMPLify \cite{Bogo:ECCV:2016}) to iteratively refine the alignment of projected 3D joints with imaged 2D joints. 
PyMAF \cite{pymaf2021} integrates such a refinement mechanism inside a feed-forward regression network, which is referred to as pyramidal mesh alignment feedback. In a similar vein, diffusion with guidance has also been adopted for the refinement \cite{Stathopoulos_2024_CVPR}.

Our focus is specifically on fully leveraging multiview observations, and we leave the use of temporal information for future work. A number of monocular temporal pose estimation methods \cite{Kanazawa_2019_CVPR, choi2020beyond, Wan_2021_ICCV, kocabas2019vibe, ye2023slahmr, wang2024tram} have also introduced inspiring mechanisms that may apply to instantaneous but multiview HMR. 
HuMoR \cite{rempe2021humor} integrates optimization in a feed-forward regression framework to ensure foot grounding and temporal consistency of the estimated pose and shape. 
TCMR \cite{choi2020beyond} extract past and future features to estimate in-between SMPL parameters.

The Transformer architecture has also been leveraged for feed-forward estimation of both non-parametric and parametric models \cite{lin2021end-to-end, you2023co, goel2023humans, Sun_2024_CVPR, multi-hmr2024, Dwivedi_2024_CVPR, Wan_2021_ICCV, Zheng_2023_CVPR, lin2023one, Qiu_2023_CVPR}.  
METRO \cite{lin2021end-to-end} predicts vertices by relating joints and vertices with a Transformer encoder.
PMCE \cite{you2023co} co-learns 3D joints and a SMPL mesh with a Transformer decoder.
HMR2.0 \cite{goel2023humans} extracts deep spatial features with a Transformer encoder and aggregates them in a SMPL query token with a Transformer decoder.
Multi-HMR~\cite{multi-hmr2024} recovers multiple people in a view with a Vision Transformer (ViT) \cite{Dosovitskiy_ICLR21} backbone.
Our work complements these works with a novel Transformer-based neural optimizer that fully leverages static multiview images. To the best of our knowledge, our method is the first neural optimizer for HMR. 

\vspace{-8pt}
\paragraph{Multiview Human Mesh Recovery}
As in any monocular vision method, single-view HMR methods are fundamentally limited by the depth ambiguity of image projection and occlusions caused by the surrounding environment, the objects the person is handling, and other body parts.
Many vision applications, including those of societal importance, let us capture the environment with multiple views. 
A handful of prior works \cite{10.1145/3652583.3658110, NEURIPS2022_33610fba, MuVS:3DV:2017} have introduced multiview methods for HMR. 
The main focus of these works are on how to aggregate multiview features pertaining to the human body shape and pose. 
Volumetric methods \cite{Chun_2023_WACV, shin2020multiviewhumanposeshape} unproject each view feature to a voxelized space similar to  multiview pose estimation \cite{iskakov2019learnable}. 
Occlusion, however, often severely impedes this aggregation and causes inaccurate mesh recovery. 
Multiview features has also been consolidated with Transformers for feed-forward SMPL parameter inference \cite{li2024humanmeshrecoveryarbitrary, Liang_2019_ICCV, jiang2022multiviewhumanbodymesh}.
Optimization-based methods \cite{huang2021dynamic, Li_2021_WACV} optimize the parameters for multiview alignment.

HeatFormer reformulates multiview HMR as a neural optimization, a seamless integration of regression and optimization, in which the Transformer encoder consolidates the heatmap observations and the decoder ``matches'' them with the current estimate of the human body and shape. Its unrolled inference corresponds to iterative refinement with image feedback and the use of heatmaps makes its end-to-end learning reliable. As a result, HeatFormer fully leverages the multiview observations regardless of their number and configuration, resolve occlusions through the novel encoder, and generalizes well to different scenes.

\section{HeatFormer}
\begin{figure*}[t]
\centering
    \includegraphics[width=\textwidth]{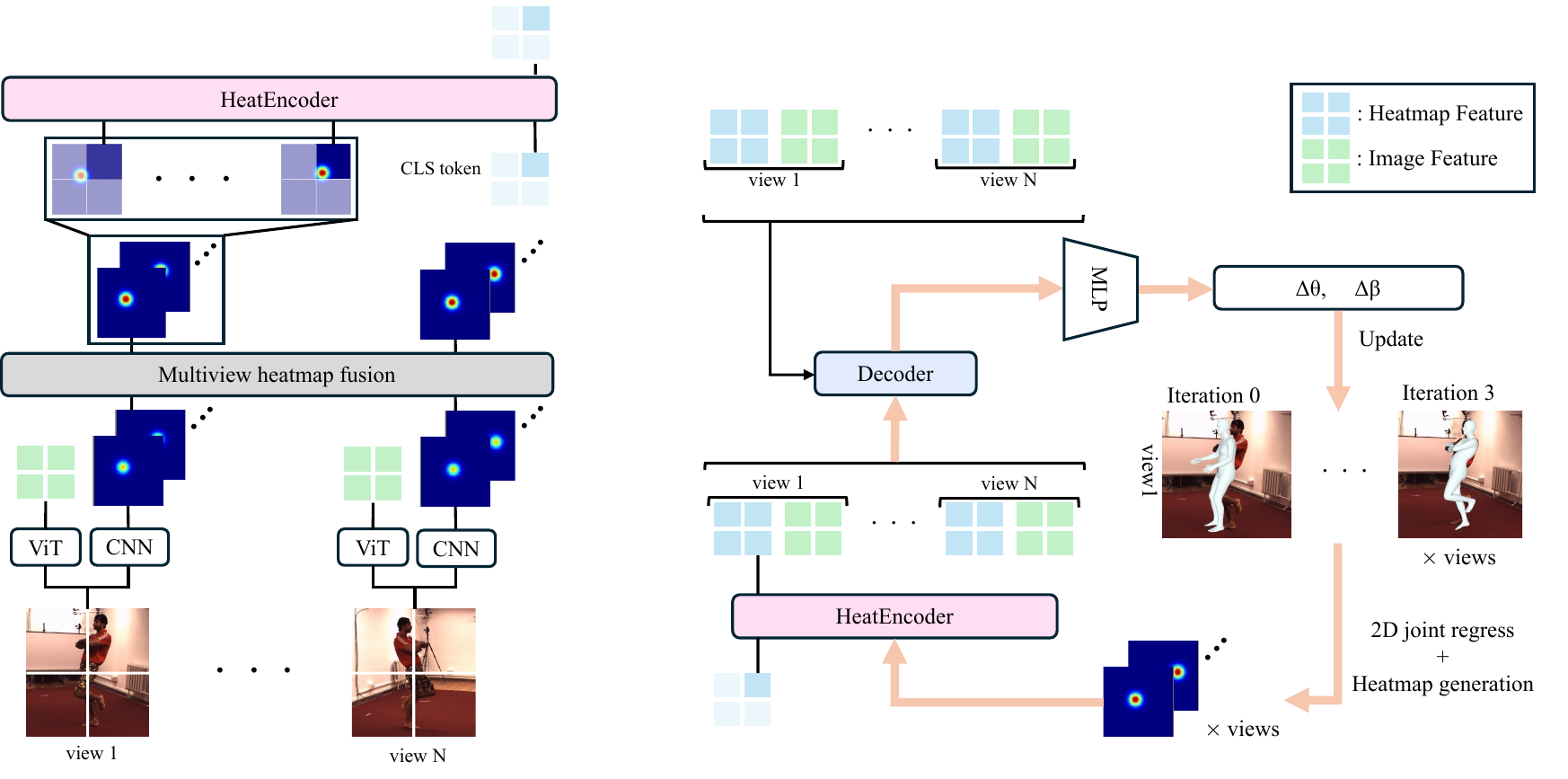}

    \caption{\vspace{-0.1\baselineskip}HeatFormer realizes neural optimization for HMR in which the Transformer encoder-decoder model serves as an unrolled iteration of SMPL fitting to the observed images. It first extracts image features and a heatmap for each view which are aggregated with a novel encoder and input to the decoder. The decoder also takes in heatmaps generated from the current SMPL estimate and, through its unrolled inference, iteratively aligns them together.}
    \label{fig:model}
\end{figure*}

We achieve occlusion-robust multiview human body shape and pose recovery with a novel transformer which we refer to as HeatFormer. \cref{fig:model} depicts the overall architecture of HeatFormer. HeatFormer realizes optimization-based HMR in a feed-forward inference in which the Transformer encoder-decoder model serves as an unrolled iteration of SMPL fitting to the observed images. A key idea underlying HeatFormer is to represent and align the body joints as heatmaps. HeatFormer forms heatmaps from the current view-dependent SMPL estimates which are iteratively aligned with the input multiview heatmaps through the encoder-decoder inference iteration. 

For this, we introduce two key components into the Transformer architecture. One is a heatmap encoder that not only better encodes the 2D joints as visual cues of the body shape and pose in the multiview observations but also realizes effective consolidation of heatmaps across the joints through self-attention.
The other is a decoder that cross-attends between the encoder-transformed multiview observations and the current body shape and pose both expressed in heatmaps. The decoder achieves this by taking in the heatmaps formed by the current SMPL estimate as queries. The encoder-decoder inference is iterated multiple times, fully leveraging the images as feedback in the neural optimization to align the SMPL heatmaps without regard to the number of views and their configuration.

\subsection{Preliminaries}

Given multiview images of a person in an environment, HeatFormer regresses the shape and pose as the parameters of a statistical human mesh model through neural optimization. 
We employ SMPL \cite{Loper:SMPL} as this statistical human body model. 
SMPL is a parametric human mesh model that captures the statistical variation of body shape and pose. Its parameters consist of the pose parameters $\theta \in \mathbb{R}^{24\times3}$ and the shape parameters $\beta \in \mathbb{R}^{10}$.
From these parameters, SMPL instantiates a human body mesh $M \in \mathbb{R}^{N\times3}$ with $N$ = 6890 vertices.
From the instantiated mesh, a joint regressor function $W \in \mathbb{R}^{N\times k}$ can return the $k$ 3D joints of the body as linear combinations of the mesh vertices. 

We derive a novel Transformer architecture that outputs these SMPL shape and pose parameters for a person captured in multiview images. The input tokens to the encoder correspond to heatmaps of the human body captured in each view. The encoder learns to consolidate heatmaps in each view which are then cross-attended with the decoder queries. The decoder learns to output a revision to the current SMPL parameter estimate and its queries are computed from the revised SMPL estimate. As a whole, HeatFormer is an unrolled iterative optimization, realized in feed-forward inference, that fits SMPL to the input views.


As the number of tokens and the spatial coordination of them do not affect the architecture, HeatFormer is fundamentally agnostic to the camera configuration. It can simply be trained to adapt to neurally optimize the SMPL model for any camera configuration, \eg, with less or more cameras. Regardless of the camera configuration, it learns to become aware and robust against the inevitable occlusions caused by real-world environments through its novel encoder. 

\subsection{HeatEncoder}

We realize HMR with a neural optimizer that integrates SMPL parameter optimization within a feed-forward network. 
This is achieved with a Transformer architecture, which iterates SMPL parameter refinement referencing the input multiview images at each unrolled feed-forward step. The multiview images carry visual evidence of the human body and pose, but in an indirect manner. A successful estimation of the SMPL parameters would align the projected body estimate to perfectly align with the person captured in the image. This alignment, however, is not easy to quantify, especially without knowledge of the true texture of the body. Most previous works rely on the 2D projected alignment of keypoints, namely body joints. These methods, employ a feed-forward regressor to output the 2D keypoints from an input image, and leverage iterative numerical optimization to align the SMPL keypoints with them.
\vspace{-0.09\baselineskip}

In contrast, our goal is to have a single consolidated feed-forward network to achieve end-to-end estimation of the SMPL parameters from the multiview inputs. For this, we need a means to quantify the alignment of the estimates with the captured person that enables robust backpropagation through the unrolled neural optimization. To achieve this, we go back to the fundamentals, namely heatmaps. A heatmap is an approximate 2D probabilistic map of the keypoint location that produces stable spatial gradients around the joint, so that gradients can be propagated stably. It was introduced exactly for this purpose, but for 2D joint localization \cite{Tompson_NIPS14,pavlakos17volumetric}. Here, our purpose is to use the heatmaps as the representation of the joints themselves throughout the neural optimization, \ie, reformulate 2D-3D alignment of body joints into a collection of 2D-2D heatmap alignments. 

To leverage heatmaps as the intrinsic representation of the body both in the multiview input images and as a surrogate of the final estimate, \ie, view-dependent SMPL parameters, we need two novel components in the Transformer architecture, a way to incorporate the body joint heatmaps from the input images, and a way to convert the output of the decoder into heatmaps so that they can be re-evaluated in the neural iteration. We achieve these by introducing a novel encoder which we refer to as HeatEncoder.

The input multiview images are first converted into image features via Vision Transformer (ViT) \cite{Dosovitskiy_ICLR21}. That is, each view is broken into patches of a regular grid, input to a pre-trained ViT, and the outputs are reassembled into a 2D image feature. We also extract heatmaps for each view. For this, we leverage AdaFuse \cite{zhang2020adafuse} which is a multiview-aware joint detector. All views are input to AdaFuse which outputs heatmaps for each view, one heatmap for each possible human joint. Now each view is converted into ViT features and multiple ($k$: the number of joints) heatmaps. We next consolidate each set of these heatmaps across different joints for each view. 

\cref{fig:model} shows the overall architecture of HeatEncoder. 
Given a set of heatmaps, $P \in \mathbb{R}^{k \times H \times W}$, where $H$ and $W$ are the heatmap height and width, respectively, corresponding to one view, HeatEncoder divides the heatmaps into patches $P_{i} \in \mathbb{R}^{k \times h \times w}$, where $i = 1,\ldots,\frac{HW}{hw}$ and $(h, w)$ are the patch size. 
Each patch is flattened and becomes a token $Q \in \mathbb{R}^{k \times \frac{H}{h} \times \frac{W}{w} \times (h \times w)}$, where $k$ is the number of tokens.

The heatmap patches are embedded with encodings that represent the joint order and the spatial position of the patch and input to the HeatEncoder as tokens together with a CLS token. HeatEncoder aggregates all these heatmaps into the CLS token via self-attention. The output CLS token is a single consolidated heatmap of all joints for that view. By repeating this HeatEncoder application to all the joint heatmaps for each view, we obtain one heatmap feature per view. For each view, we concatenate each patch of the consolidated heatmap with the image feature patch at the corresponding position to form a token (\ie, we obtain $\frac{H}{h} \times \frac{W}{w}$ tokens for the $N$ views). 

We pretrain this HeatEncoder without the image features by appending a CLS token in the input and directly regressing the SMPL parameters from its output.

\begin{figure}
  \centering
  \includegraphics[width=\linewidth]{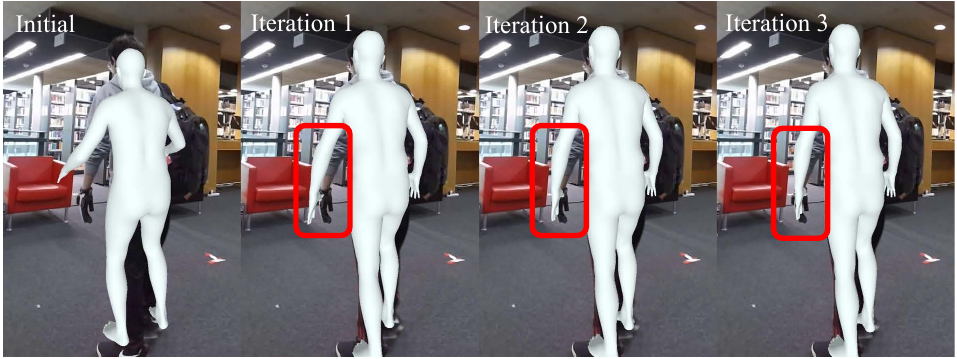}
  \caption{HeatFormer is an unrolled iterative optimizer realized through its forward inference. HeatFormer converges to accurate SMPL estimates within three unrolled inferences.}
  \label{fig:optimization_step}
\end{figure}

\subsection{Decoder As A Neural Optimizer}

We realize neural optimization with the Transformer architecture by repeated inference in the decoder. Multiple feed-forward passes through the decoder corresponds to unrolled iteration of an optimization. The SMPL estimates are output multiple times through these inferences which refines the estimates. The heatmaps generated from the current estimates (query) are cross-attended with those of the input multiview images (key and value), both of which are consolidated with the HeatEncoder, leading to their alignment. 

We output view-dependent human shape and pose so that our model is agnostic to the number of views. These outputs can be consolidated into a single SMPL parameter in 3D space or by simply picking the best estimate in terms of 2D overlap, or even just selecting the one closest to the view used in downstream applications. This flexibility is also an advantage of recovering view-dependent estimates.

We convert the estimated view-dependent SMPL parameters into heatmaps so that they can be iteratively aligned with those of the input views with unrolled inference through the decoder. For each view, we first instantiate the SMPL mesh and recover their joints with the joint regressor. We then draw Gaussians centered around each joint to obtain heatmaps for the $k$ joints. These are then patchified as tokens and input to the HeatEncoder, in the same manner as the input views, to extract consolidated heatmaps. This consolidated heatmap is then concatenated with the same ViT features for each view and input to the decoder as queries. We reuse corresponding ViT features from the input images so that associations between the same view is stronger in the decoder cross-attention. 

The outputs of the decoder are reshaped into $N \times C \times (h \times w)$ tensors, where $N$ is the number of views and $C$ is the channel size, and passed through average pooling and an MLP, so that they represent SMPL parameter revisions to the current SMPL estimates, \ie, the difference that should be added to the current estimate. The SMPL parameters are revised and new heatmaps for each view are generated from SMPL instantiated for each view. These heatmaps are used as new queries of the decoder and the revision is computed through cross-attention again. This repeated forward inference realizes neural optimization. In our experiments, we used three iterations, \ie, unrolled forward inferences. As shown with an example in \cref{fig:optimization_step}, we observed clear convergence for all cases within these three or four iterations.

For the initial estimates, we use the SMPL mean parameters for the pose and shape, following previous works such as HMR~\cite{kanazawaHMR18}.
The use of mean parameters when computing the global orientation of the body can cause erroneous alignment (\eg, a forced twist in the body). For this, we estimate the global orientation in advance for each view individually with a pre-trained Transformer.
The translation can be easily recovered when the cameras are externally calibrated. 
HeatFormer can be also applied without camera calibration. The camera extrinsics are only used for opportunistic sanitization of heatmaps with AdaFuse and query heatmap projection. These steps can be either skipped or instead estimated, respectively, when the camera extrinsics are unknown. For the latter, when using uncalibrated cameras, we can use the weak-perspective camera parameters regressed from image features for each view similar to HMR2.0~\cite{goel2023humans}.  Please see the appendix for further details.

\subsection{Loss Function And Training}
Following prior work on HMR, we train our model using losses on the 3D joints, 2D joints, and SMPL parameters, and leverage an adversarial loss. We also add losses on the heatmap and mesh vertices.
Our model predicts SMPL parameters $(\theta, \beta)$ and the weak-perspective camera parameters given input multiview images.
The 3D joint loss $\mathcal{L}_\textrm{3D}$ is computed as the mean-squared error (MSE) of the joints recovered from the estimated SMPL mesh from the ground-truth 3D joints. 
The 2D joint loss $\mathcal{L}_\textrm{2D}$ is computed as the MSE of the view-projected 3D joint estimates from the cropped and normalized ground-truth 2D joints.

The SMPL parameter loss $\mathcal{L}_\textrm{smpl}$ is computed as the MSE between the SMPL estimates and their annotated (pseudo ground-truth) parameter values in the dataset. 

To encourage the decoder inference to learn the process of aligning heatmaps, we introduce a step-weighted heatmap loss. This step-weighted heatmap loss measures the discrepancy between the heatmaps weighted by a scalar that gradually increases as the iteration progresses 
\begin{equation}
\mathcal{L}_\textrm{hm} = \sum_{i}{\lambda_{i}\|M^{\dag} - M_{i}\|}\,,
\end{equation}
where \dag\ represents pseudo ground-truth  and $i$ indexes the iteration number, $\lambda_{i}$ is the loss weight and $M$ is a heatmap. Note again that we typically repeat the forward inference 3 times ($i=1, 2, 3$). 

Sometimes the body would try to twist to align the 2D heatmaps. We avoid this with a vertex loss that measures the MSE of 3D vertices on the SMPL mesh instantiated with the estimated and dataset-annotated SMPL parameters. This loss is computed for each repeated decoder inference. 

We follow HMR~\cite{kanazawaHMR18} and employ an adversarial prior and compute an adversarial loss $\mathcal{L}_{adv}$ to discourage unrealistic human poses. The complete loss becomes 
\begin{equation}
\scalebox{0.85}{$
\mathcal{L} = \lambda_\textrm{3D}\mathcal{L}_\textrm{3D} + \lambda_\textrm{2D}\mathcal{L}_\textrm{2D} + \mathcal{L}_\textrm{smpl} + \mathcal{L}_\textrm{hm} + \lambda_\textrm{v}\mathcal{L}_{\textrm{v}} + \lambda_\textrm{adv}\mathcal{L}_\textrm{adv}\,,
$}
\end{equation}
where $\mathcal{L}_\textrm{smpl}$ is a weighted sum of MSE on the pose $\theta$ and shape $\beta$ parameters with corresponding weights.

We train HeatFormer with a batch size of 8, typically 4 views, a learning rate of $1 \times 10^{-5}$,  repeated inference of 3 times for 50 epochs, and set the loss weights to $\lambda_\textrm{3D} = 0.05, \lambda_\textrm{2D} = 0.01, \lambda_{\theta} = 0.001, \lambda_{\beta} = 0.0005, \lambda_{\textrm{v}} = 0.0003, \lambda_\textrm{hm} = [0.001, 0.003, 0.005], \lambda_\textrm{adv} = 0.0005$.
The HeatEncoder consists of 2 layers and 8 heads and the CLS token dimension is 1024.
The decoder consists of 2 layers and 8 heads.
In our model, input image size is 384 for heatmap extraction and 256 for Transformer architecture.

Note that HeatFormer is agnostic to the number of views and camera configuration. It can be trained on a number of views different from the number of cameras used at inference time. The decoder iteration can also be varied, but we empirically found that the neural optimization converges in 3 or 4 iterations. We experimentally validate all these. 

\section{Experiments}

We evaluate HeatFormer by focusing on its three key aspects. First is its accuracy. We quantitatively evaluate its accuracy on major datasets and compare the results with state-of-the-art methods. Second is its generalization capability. We apply HeatFormer to datasets that are not included in training and show that it can estimate human pose and shape in a variety of environments. Third is its occlusion robustness. We evaluate the accuracy on the BEHAVE dataset with natural object occlusion. We also conduct thorough ablation studies to validate the effectiveness of our model components and compare the accuracy for varying numbers of views.

\vspace{-8pt}
\paragraph{Datasets and Metrics}
We pre-train HeatEncoder on Human3.6M \cite{6682899} and MPI-INF-3DHP \cite{mono-3dhp2017} with single-view. 
We then freeze HeatEncoder and ViT and train the entire HeatFormer on Human3.6M and MPI-INF-3DHP with multiview images. 
Following previous work \cite{Liang_2019_ICCV, shin2020multiviewhumanposeshape, jia2023paff, li2024humanmeshrecoveryarbitrary} on multiview HMR, we train on subjects 1, 5, 6, 7, and 8 and test on subjects 9 and 11 for Human3.6M and train on subjects 1 to 7 and test on subject 8 for MPI-INF-3DHP.
We use the metrics MPJPE, PA-MPJPE, and MPVPE to evaluate for Human3.6M dataset.
MPJPE/MPVPE is the Mean Per Joint/Vertex Position Error and PA means procrustes aligned.
For the MPI-INF-3DHP dataset, we use the metrics MPJPE, PCK, and AUC.
Note that we evaluate all metrics after procrustes alignment for MPI-INF-3DHP.
PCK (Percentage of Correct Keypoints) represents the percentage of keypoints less than or equal to a predetermined distance from the ground truth. We set the threshold to 150mm following previous works.
AUC (Area Under the Curve) represents the area under the PCK curve when changing the PCK threshold in the range of 0-150mm in 5mm increments.

\vspace{-8pt}
\paragraph{Accuracy}
\cref{tab:performance_Human3.6M} shows the accuracy of HeatFormer in comparison with state-of-the-art methods for both single-view and multiview methods evaluated on the Human3.6M dataset. The results, of course except ours, are those reported in their respective papers to ensure fairness. All methods are trained on their own combinations of datasets. Our HeatFormer was trained on the training splits of Human3.6M and MPI-INF-3DHP. Ours\dag \ uses ground-truth 3D joints to estimate the translation. It is shown as reference to show the effect of translational errors. 
HeatFormer achieves SOTA accuracy and demonstrates the effectiveness of fully leveraging multiview images with a neural optimizer. 

To confirm the effectiveness of neural optimization, we evaluate the inference results after one, two, and three inference steps through the decoder of a 3-iter model. The results shown in \cref{tab:performance_iter} show that our model quickly converges.
\cref{fig:results} shows visual comparisons. Please see the appendix for more qualitative results.

\begin{table}
  \centering
  \small{
  \begin{tabular}{@{}lccc@{}}
    \toprule
    \textbf{Method} & \textbf{MPJPE $\downarrow$} & \textbf{PA-MPJPE $\downarrow$} \\
    \hline
    HMR2.0(a) \cite{goel2023humans} & 44.8 & 33.6\\
    HMR2.0(b) \cite{goel2023humans} & 50.0 & 32.4\\
    HMR2.0+scoreHMR-a \cite{Stathopoulos_2024_CVPR} & 47.9 & 28.4\\
    HMR2.0+scoreHMR-b \cite{Stathopoulos_2024_CVPR} & 44.7 & 29.0 \\
    PostureHMR \cite{Song_2024_CVPR} & 44.5 & 31.0\\
    \hline
    Shape-Aware \cite{Liang_2019_ICCV} & 79.9 & 45.1\\
    MV-SPIN \cite{shin2020multiviewhumanposeshape} & 49.8 & 35.4\\
    LVS \cite{shin2020multiviewhumanposeshape} & 46.9 & 32.5\\
    ProHMR \cite{kolotouros2021prohmr} & 62.2 & 34.5\\
    Yu et al. \cite{NEURIPS2022_33610fba} & - & 33.0 \\
    PaFF \cite{jia2023paff} & 33.0 & 26.9\\
    U-HMR \cite{li2024humanmeshrecoveryarbitrary} & \cellcolor{yellow!25}31.0 & \cellcolor{orange!25}22.8\\
    \hline
    \textbf{HeatFormer (Ours iter3)} & \cellcolor{orange!25}30.7 & \cellcolor{yellow!25}23.3\\
    \textbf{HeatFormer (Ours iter4)} & \cellcolor{red!25}29.5 & \cellcolor{red!25}22.4\\
    Reference (Ours\dag\ iter3) & 28.6 & 23.1\\
    \bottomrule
  \end{tabular}}
  \caption{Accuracy on Human3.6M. Our method achieves SOTA of published works.}
  \label{tab:performance_Human3.6M}
\end{table}

\begin{table}
  \centering
  \small{
  \begin{tabular}{@{}l|cccc@{}}
    \toprule
    \textbf{Method} & \textbf{MPJPE $\downarrow$} & \textbf{PCK $\uparrow$} & \textbf{AUC $\uparrow$} \\
    \hline
    Shape-Aware \cite{Liang_2019_ICCV} & 59.0 & 95.0 & 65.0 \\
    LVS \cite{shin2020multiviewhumanposeshape} & 50.2 & 97.4 & 65.0 \\
    PaFF \cite{jia2023paff} & \cellcolor{yellow!25}48.4 & \cellcolor{yellow!25}98.6 & \cellcolor{yellow!25}67.3 \\
    \hline
    \textbf{HeatFormer (Ours iter3)} & \cellcolor{red!25}39.8 & \cellcolor{red!25}99.5 & \cellcolor{red!25}72.8 \\
    \textbf{HeatFormer (Ours iter4)} & \cellcolor{orange!25}40.6 & \cellcolor{orange!25}99.5 & \cellcolor{orange!25}72.3\\
    Reference (Ours\dag\ iter3) & 36.0 & 99.5 & 75.3 \\
    \bottomrule
  \end{tabular}}
  \caption{Evaluation on MPI-INF-3DHP. HeatFormer generalizes across different scenes and camera configurations thanks to its neural optimization formulation.}
  \label{tab:performance_mpii3d}
\end{table}


\begin{table}
  \centering
  \scalebox{0.7}{
  \begin{tabular}{@{}l|ccc|cccc@{}}
    \toprule
      & \multicolumn{3}{c|}{\textbf{Human3.6M}} & \multicolumn{3}{c}{\textbf{MPI-INF-3DHP}} \\
    \textbf{Method} & \textbf{MPJPE $\downarrow$} & \textbf{PA-MPJPE $\downarrow$} & \textbf{MPVPE $\downarrow$} & \textbf{MPJPE $\downarrow$} & \textbf{PCK $\uparrow$} & \textbf{AUC $\uparrow$}\\
    \hline
    iter1 & \cellcolor{yellow!25}34.9 & \cellcolor{yellow!25}26.2 & \cellcolor{yellow!25}41.9 & \cellcolor{yellow!25}44.6 & \cellcolor{orange!25}99.3 & \cellcolor{yellow!25}70.0 \\
    iter2 & \cellcolor{orange!25}31.2 & \cellcolor{orange!25}23.6 & \cellcolor{orange!25}37.5 & \cellcolor{orange!25}40.8 & \cellcolor{red!25}99.5 & \cellcolor{orange!25}72.1 \\
    iter3 & \cellcolor{red!25}30.7 & \cellcolor{red!25}23.3 & \cellcolor{red!25}37.0 & \cellcolor{red!25}39.8 & \cellcolor{red!25}99.5 & \cellcolor{red!25}72.8 \\
    \bottomrule
  \end{tabular}}
  \caption{Accuracy for each forward inference (\ie, optimization iteration). The accuracy steadily improves and converges within three forward inferences for both datasets.}
  \label{tab:performance_iter}
\end{table}



\vspace{-8pt}
\paragraph{Generalization}
To evaluate how well the method generalizes to different scenes and view configurations, we train and test on unseen scenes and also altogether different datasets. 
We first conduct a cross-dataset evaluation by training on the Human3.6M dataset and testing on the MPI-INF-3DHP dataset. We compare with the multiview method that achieved high-accuracy on Human3.6M, U-HMR~\cite{li2024humanmeshrecoveryarbitrary}. 
As shown in  \cref{tab:cross-mpiinf3dhp}, our HeatFormer achieves higher accuracy across all metrics. These results show that our HeatFormer generalizes well to unseen scenes and camera configurations thanks to its neural optimization formulation. On the other hand, the results show that U-HMR is likely overfit to Human3.6M. We also evaluate the accuracy of HeatFormer trained on Human3.6M and MPI-INF-3DHP \cite{6682899, mono-3dhp2017} when applied to the BEHAVE dataset in our results for occlusion evaluation. These results also demonstrate the strong generalizability of our method.

\begin{table}
  \centering
  \scalebox{0.8}{
  \begin{tabular}{@{}l|cccc@{}}
    \toprule
    \textbf{Method} & \textbf{MPJPE $\downarrow$} & \textbf{PCK $\uparrow$} & \textbf{AUC $\uparrow$} \\
    \hline
    U-HMR \cite{li2024humanmeshrecoveryarbitrary} & 73.2 & 91.8 & 52.5 \\
    \hline
    \textbf{HeatFormer (Ours iter3)} & \cellcolor{red!25}{56.0} & \cellcolor{red!25}{96.9} & \cellcolor{red!25}{62.7} \\
    Reference (Ours\dag\ iter3) & 55.4 & 96.8 & 63.1 \\
    \bottomrule
  \end{tabular}}
  \caption{Cross-dataset evaluation. We trained on Human3.6M and evaluate on MPI-INF-3DHP. HeatFormer generalizes well across different types of scenes.}
  \label{tab:cross-mpiinf3dhp}
\end{table}

\vspace{-8pt}
\paragraph{Occlusion}
HeatFormer is robust to occlusions by design as it can fully leverage multiview images through the input tokens, the heatmap consolidation with HeatEncoder, and the heatmap alignment with unrolled decoder inference. We evaluate this robustness to occlusion using the BEHAVE dataset \cite{bhatnagar22behave} with object occlusions.
In this dataset, we use the 15 body joints except the eye, ear and toe joints of the 25 body joints in openpose \cite{8765346}.
We evaluate the accuracy against the occlusions by objects the person handles and also other body parts with 2 different protocols.
One is the evaluation of all joints, the other is the evaluation of joints which have confidence scores higher than $0.3$. These confidence scores are given in the dataset. 
About 70\% of all joints has confidence above $0.3$.
We compare the occlusion robustness and also cross-dataset generalization power on the BEHAVE dataset with HMR2.0 \cite{goel2023humans} as it is a monocular method particularly claimed to be robust to occlusions and also pre-trained on a large set of datasets and multiview refinement methods \cite{kolotouros2021prohmr, Stathopoulos_2024_CVPR}. 
Our HeatFormer is trained on the Human3.6M and MPI-INF-3DHP datasets, and HMR2.0 is trained on the Human3.6M, MPI-INF3DHP, COCO \cite{lin2014microsoft} and MPII \cite{Andriluka_2014_CVPR} datasets. 
As the results in \cref{tab:cross-behave} clearly show, HeatFormer achieves high accuracy for both protocols, which demonstrates its robustness to occlusion and also its strength in generalization to different scenes and view configurations. 

\begin{table}
  \centering
  \scalebox{0.7}{
  \begin{tabular}{@{}l|cc|ccc@{}}
    \toprule
     & \multicolumn{2}{c|}{\textbf{Protocol1}} & \multicolumn{2}{c}{\textbf{Protocol2}} \\
    \textbf{Method} & \textbf{MPJPE$\downarrow$} & \textbf{PA-MPJPE$\downarrow$} & \textbf{MPJPE$\downarrow$} & \textbf{PA-MPJPE$\downarrow$} \\
    \hline
    HMR2.0a \cite{goel2023humans} & \cellcolor{yellow!25}72.2 & 41.2 & \cellcolor{yellow!25}48.1 & 26.9 \\
    HMR2.0b \cite{goel2023humans} & 94.0 & 53.1 & 63.2 & 34.7 \\
    \hline
    ProHMR \cite{kolotouros2021prohmr} & 111.2 & 55.7 & 74.7 & 37.1\\
    HMR2.0a+scoreHMR \cite{Stathopoulos_2024_CVPR} & 72.9 & \cellcolor{red!25}30.9 & 49.0 & \cellcolor{red!25}20.1\\
    HMR2.0b+ScoreHMR \cite{Stathopoulos_2024_CVPR} & 86.0 & 35.3 & 58.0 & 22.8\\
    \hline
    \textbf{HeatFormer (Ours iter3)}& \cellcolor{orange!25} 51.1 & \cellcolor{yellow!25} 32.2 & \cellcolor{orange!25}34.2 & \cellcolor{yellow!25}21.2 \\
    \textbf{HeatFormer (Ours iter4)} & \cellcolor{red!25} 48.9 & \cellcolor{orange!25} 31.8 & \cellcolor{red!25} 32.6 & \cellcolor{orange!25} 20.9 \\
    Reference (Ours\dag\ iter3) & 51.8 & 33.0 & 34.7 & 21.7 \\
    \bottomrule
  \end{tabular}}
  \caption{Accuracy against occlusions evaluated on the BEHAVE dataset and cross-dataset evaluation. See text for Protocol 1 and 2. Our HeatFormer is considerably more robust to occlusions than large-scaled monocular methods, thanks to its ability to fully leverage multiview images.}
  \label{tab:cross-behave}
\end{table}


\vspace{-8pt}
\paragraph{View Configuration}
HeatFormer is fundamentally a Transformer and can be trained and inferred on a different number of tokens, \ie, views. We demonstrate this by training HeatFormer on fixed 1, 2, 3, 4 views from the Human3.6M dataset as separate models and test it using those numbers of views.
As can be seen in the results shown in \cref{tab:performance_view}, the accuracy increases as the number of views increases.
Please note that, since for calibrated cameras an accurate translation cannot be computed for single-view, we use the calibration-free model only for the single-view.

\cref{tab:performance_view_all} shows results on training and testing on different numbers of views. The architecture of HeatFormer allows discrepancy in the number of views used for training and testing and still maintains high accuracy. This flexibility in view configurations, thanks to the Transformer-based neural optimization formulation, is a key strength that likely significantly adds to the practicality of HeatFormer.

Please see the appendix for more results including ablation studies on the HeatEncoder.

\begin{table}
    \centering
  \scalebox{0.8}{
    \begin{tabular}{c|cccc}
    \toprule
     & \multicolumn{4}{c}{Number of Views} \\
     & 1 & 2 & 3 & 4\\
    \hline
     MPJPE$\downarrow$ & 47.1 & \cellcolor{yellow!25}35.2 & \cellcolor{orange!25}32.0 & \cellcolor{red!25}29.3 \\
     PA-MPJPE$\downarrow$ & 34.3 & \cellcolor{yellow!25}27.9 & \cellcolor{orange!25}25.2 & \cellcolor{red!25}23.2 \\
     MPVPE$\downarrow$ & 60.8 & \cellcolor{yellow!25}45.4 & \cellcolor{orange!25}41.1 & \cellcolor{red!25}36.7 \\
    \bottomrule
    \end{tabular}}
    \caption{The effect of using different numbers of views evaluated on Human3.6M. We use the same number of views for training and testing. The accuracy increases with more views. This view configuration flexibility is a key advantage of HeatFormer.}
    \label{tab:performance_view}
\end{table}


\begin{table}
    \centering
  \scalebox{0.8}{
    \begin{tabular}{lc|cccc}
    \toprule
      & & \multicolumn{4}{c}{Testing \# of Views} \\
       & & 1 & 2 & 3 & 4\\
    \hline
    \multirow{4}{*}{Training \# of Views} 
            & 1 & \cellcolor{red!25}34.3 & \cellcolor{yellow!25}36.0 & \cellcolor{yellow!25}36.0 & \cellcolor{orange!25}34.5 \\
            & 2 & 46.5 & \cellcolor{orange!25}27.9 & \cellcolor{yellow!25}28.5 & \cellcolor{red!25}27.0 \\
            & 3 & 45.2 & \cellcolor{yellow!25}31.3 & \cellcolor{red!25}25.2 & \cellcolor{orange!25}25.5 \\
            & 4 & 47.0 & \cellcolor{yellow!25}31.1 & \cellcolor{orange!25}26.2 & \cellcolor{red!25}23.2 \\
    \bottomrule
    \end{tabular}}
    \caption{Results for different combinations of different numbers of training and testing views (Human3.6M. Metric is PA-MPJPE). HeatFormer's Transformer architecture makes it fundamentally agnostic to the number of views, and they can even differ between training and testing while maintaining accuracy. Note that the diagonal (same \# views) would tend to give the highest accuracy.}
    \label{tab:performance_view_all}
\end{table}

\section{Conclusion}
We introduced HeatFormer, a novel neural human body shape and pose optimizer. HeatFormer fully leverages multiview images with its HeatEncoder that consolidates joint heatmaps and transforms SMPL estimation into heatmap alignment through unrolled iterative inference through the decoder. Experimental results demonstrate the effectiveness of HeatFormer over other monocular and multiview methods and show that it is accurate, robust to occlusions, generalizes well, and adapts to different view configurations. Human behavior understanding from multiview cameras fixed to an environment can benefit a wide range of applications in vision, robotics, security, commerce, and well-being. We believe that HeatFormer would serve as an effective foundational tool for this. 

\begin{figure}
  \centering
  \includegraphics[width=\linewidth]{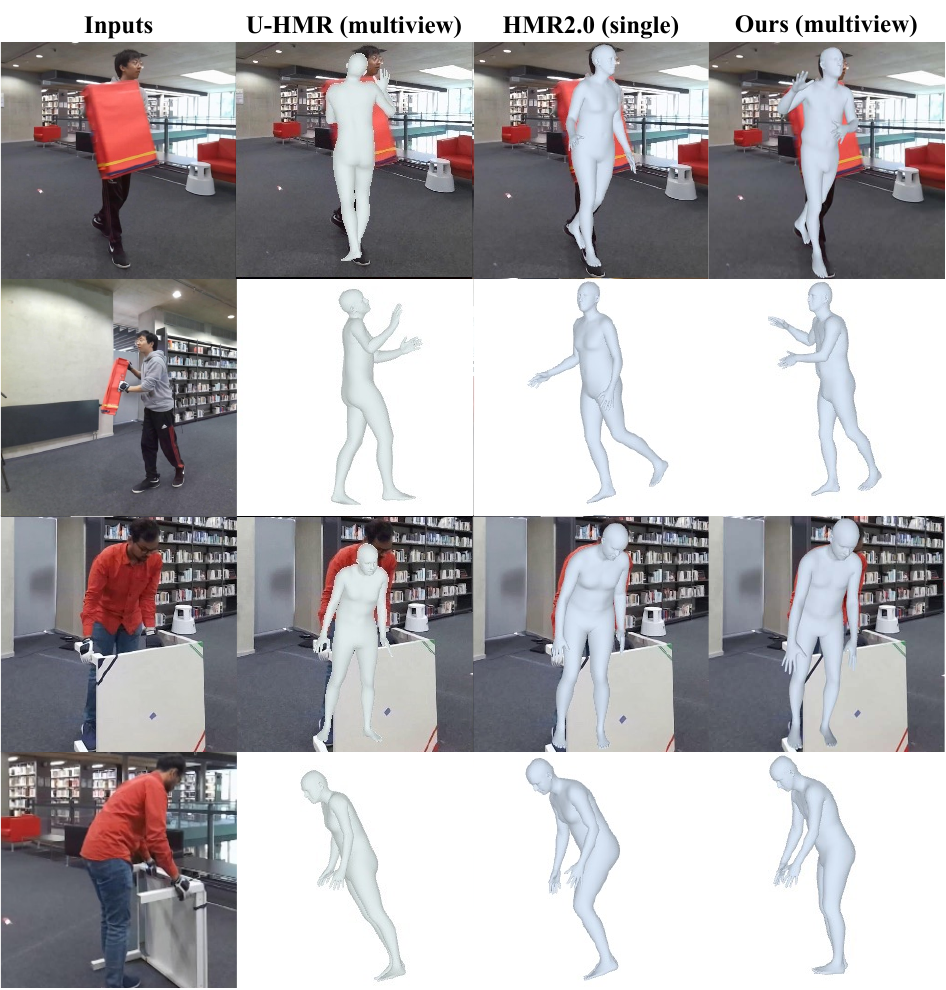}
  \caption{HeatFormer can reconstruct occluded body parts by referencing other views even though they are capturing the occluded body part from a different viewing direction. This complementary use of multiview images is the core strength of HeatFormer, which complements monocular methods like HMR2.0.}
  \label{fig:results}
\end{figure}

\section*{Acknowledgements}
This work was in part supported by
JSPS 
20H05951 and 
21H04893, 
JST JPMJCR20G7 
and JPMJAP2305, 
and RIKEN GRP.

\appendix

\setcounter{figure}{0}
\setcounter{table}{0}

\renewcommand\thefigure{\Alph{figure}}
\renewcommand\thetable{\Alph{table}}
\renewcommand{\dbltopfraction}{0.99}
\renewcommand{\textfraction}{0.01}
\renewcommand{\floatpagefraction}{0.99}
\renewcommand{\dblfloatpagefraction}{0.99}

\section{Implementation Details}
\paragraph{HeatEncoder}
We train HeatEncoder using 1 NVIDIA RTX A6000 with a batch size of 32 and use the AdamW optimizer with a learning rate of 1e-5 for 50 epochs. The learning rate is multiplied by 0.2 each time it reaches 20, 30, and 40 epochs. We use the 3D joint loss, 2D joint loss, and SMPL parameter loss and train on Human3.6M \cite{6682899} and MPI-INF-3DHP \cite{mono-3dhp2017} datasets for about 3.5K iterations per epoch. The training data ratio is approximately 2:1. HeatEncoder training takes about three days.

\paragraph{HeatFormer}
We then freeze HeatEncoder and train the entire HeatFormer using 1 NVIDIA A100 with a batch size of 8 and 4 views for each batch. Same as the HeatEncoder, we use the AdamW optimizer with a learning rate of 1e-5 for 50 epochs and the learning rate is multiplied by 0.2 each time it reaches 30 and 40 epochs. The training dataset ratio is similar to HeatEncoder.
HeatFormer training takes about six days.

\section{Dataset Details}
We describe the details of Human3.6M \cite {6682899}, MPI-INF-3DHP \cite{mono-3dhp2017}, BEHAVE \cite{bhatnagar22behave} and RICH \cite{Huang:CVPR:2022} datasets.
The BEHAVE and RICH datasets are used only for testing.

\paragraph{Human3.6M}
We preprocess the Human3.6M dataset following \cite{choi2020beyond}.
Human3.6M does not have ground-truth SMPL parameters. Instead, we use the pseudo ground-truth SMPL parameters generated by NeuralAnnot \cite{Moon_2022_CVPRW_NeuralAnnot}.
We sample the dataset every 20 frames which amounts to about 20K frames of training data for each view.

\paragraph{MPI-INF-3DHP}
Same as the Human3.6M dataset, we preprocess the data following \cite{choi2020beyond} and use pseudo ground-truth SMPL parameters generated from NeuralAnnot \cite{Moon_2022_CVPRW_NeuralAnnot}.
We removed data whose MPJPE computed on the pseudo ground-truth SMPL parameters exceeds 40mm as they lack reliable ground truth on only train split.
We sample every 10 frames which results in about 10K frames for each view.

\paragraph{BEHAVE}
The BEHAVE dataset is a dataset capturing, with 4 views, human-object interactions in natural environments.
We use the BEHAVE dataset to evaluate the geenralization capability and occlusion-robustness of our model.
We follow the train and test splits of the BEHAVE dataset and evaluate and compare on the test data.
Qualitative results on the BEHAVE dataset are shown in \cref{sec:qual}.

\paragraph{RICH}
The RICH dataset is a real scene dataset taken from 4 views.
We show qualitative results on the RICH dataset in \cref{sec:qual}.

\section{Qualitative Results}
\label{sec:qual}
\begin{figure*}[t]
\centering
    \includegraphics[width=\textwidth]{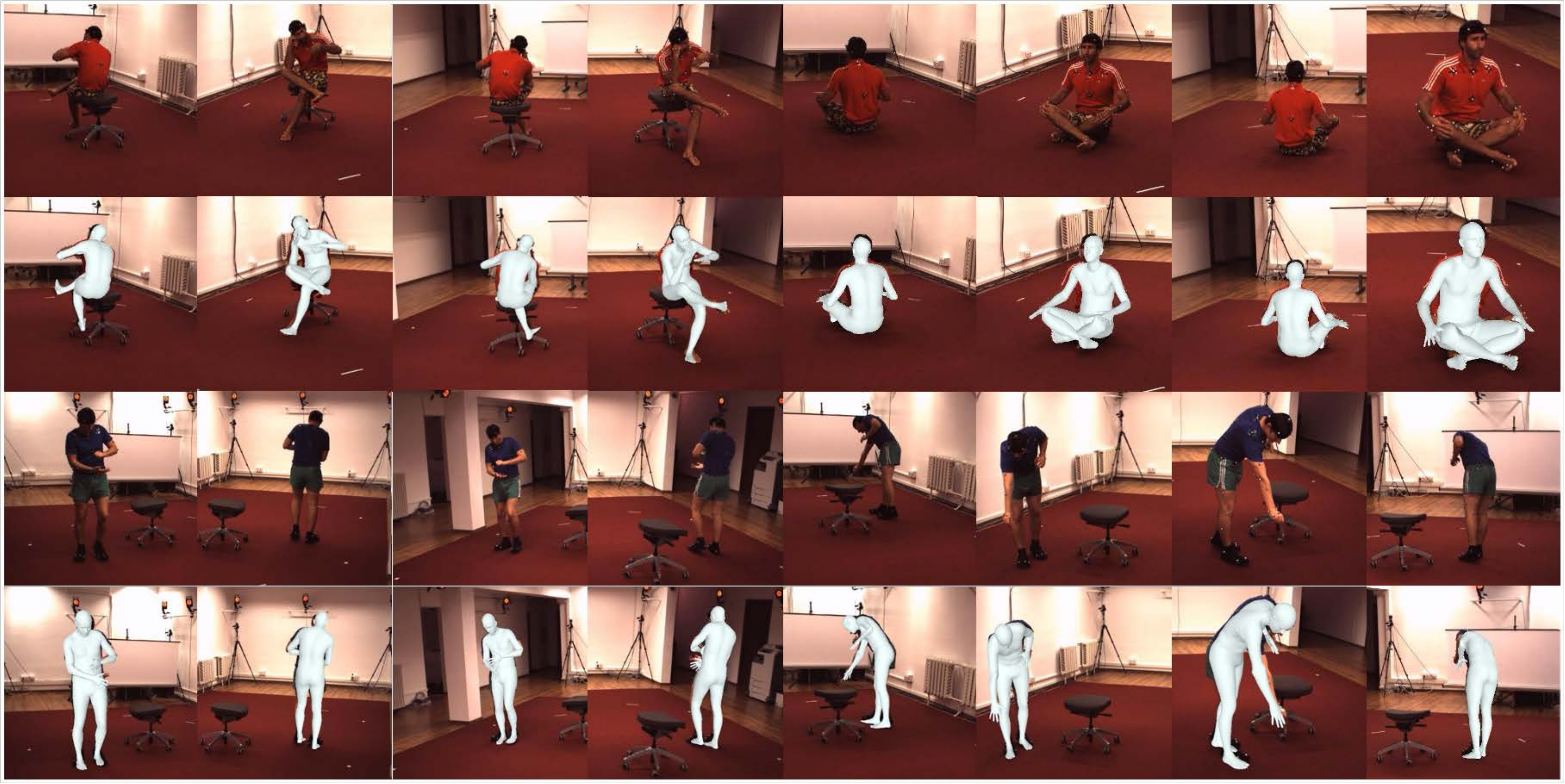}
    \caption{Qualitative results on the Human3.6M \cite{6682899} dataset. HeatFormer successfully leverages the multiview observations to resolve the complex occlusions.}
    \label{fig:Human3.6M}
\end{figure*}

\begin{figure*}[t]
\centering
    \includegraphics[width=\textwidth]{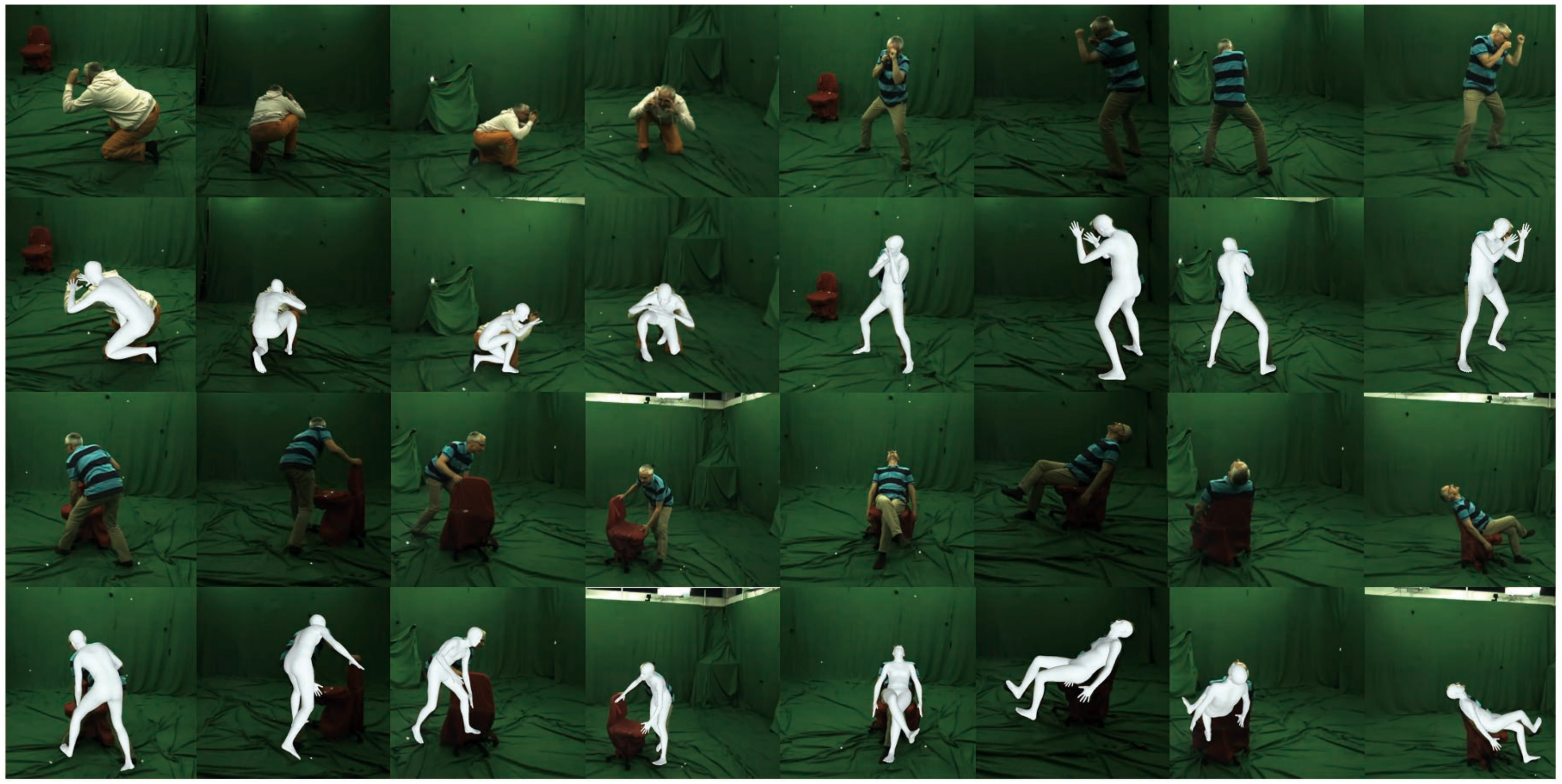}
    \caption{Qualitative results on the MPI-INF-3DHP \cite{mono-3dhp2017} dataset. The body shape and pose behind various kinds of occlusions are successfully recovered.}
    \label{fig:MPI-INF-3DHP}
\end{figure*}

\begin{figure*}[t]
\centering
    \includegraphics[width=\textwidth]{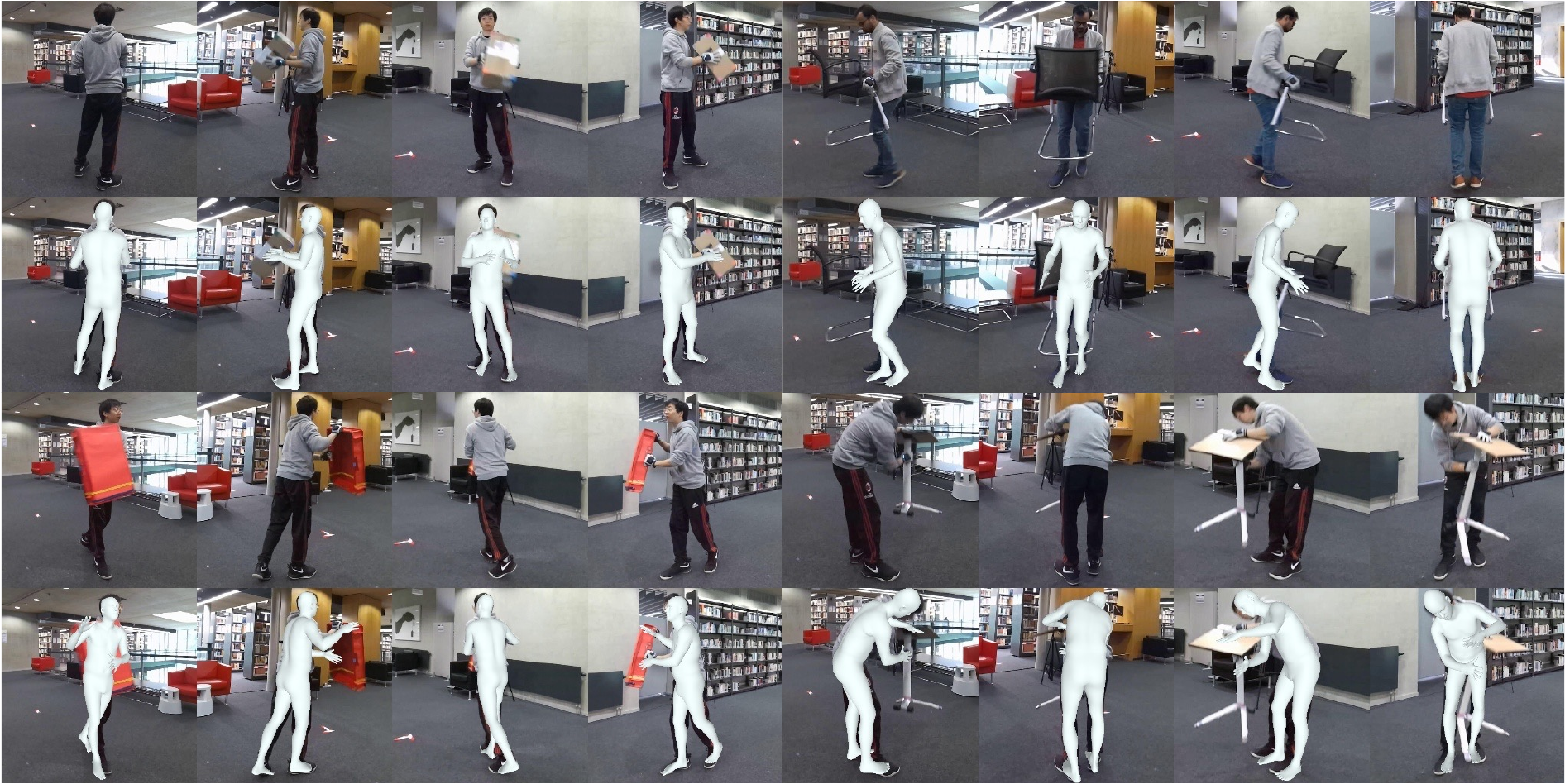}
    \caption{Qualitative results on the BEHAVE \cite{bhatnagar22behave} dataset. This dataset is not used in training. HeatFormer generalizes well to unseen scenes and unseen types of occlusion.}
    \label{fig:BEHAVE}
\end{figure*}

\begin{figure*}[t]
\centering
    \includegraphics[width=\textwidth]{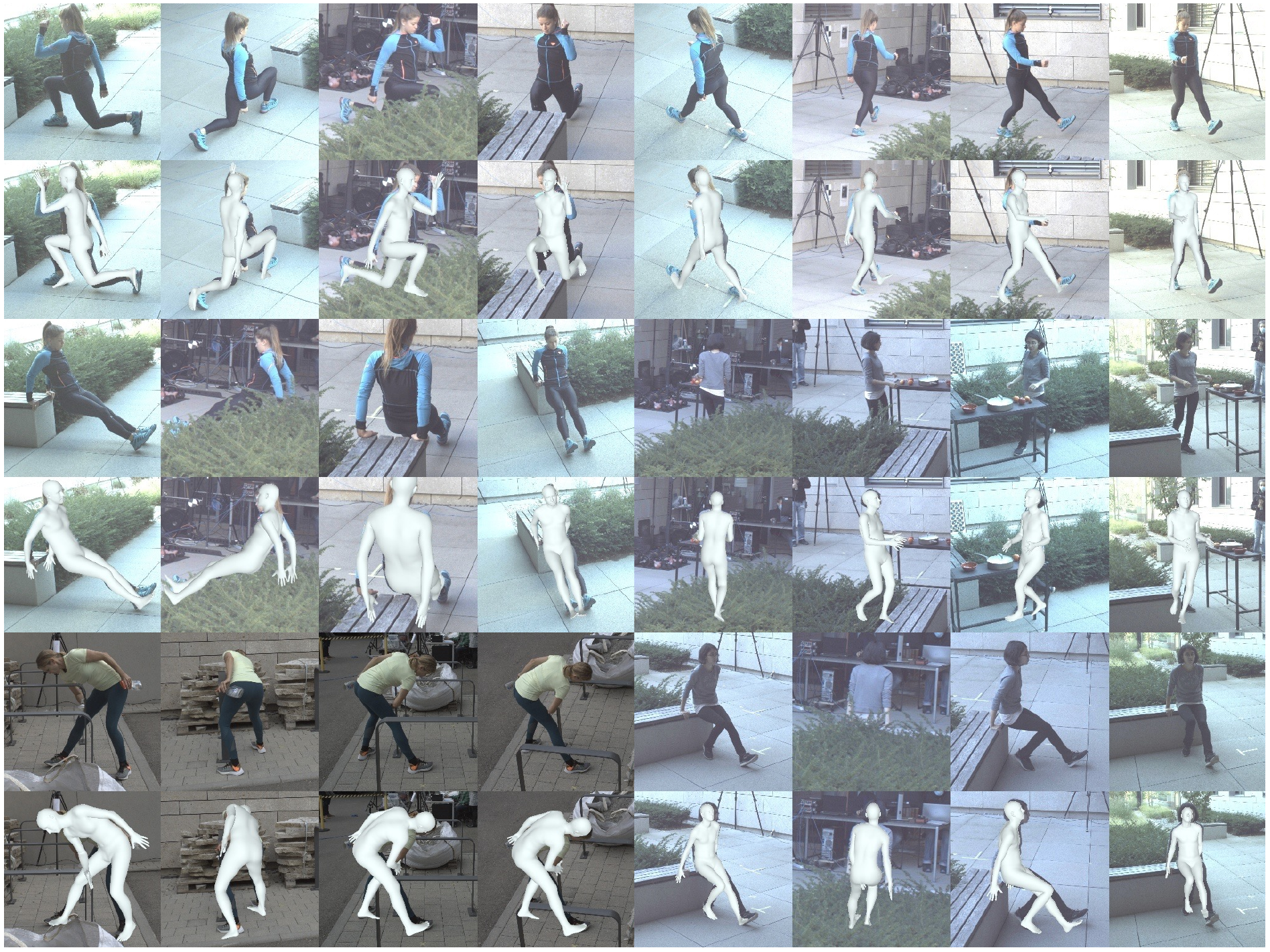}
    \caption{Qualitative results on RICH \cite{Huang:CVPR:2022} dataset. This dataset is not used in training. The results clearly demonstrate the strong generalization capability and occlusion-robustness of HeatFormer.}
    \label{fig:RICH}
\end{figure*}

We show qualitative results for different datasets, Human3.6M \cref{fig:Human3.6M}, MPI-INF-3DHP \cref{fig:MPI-INF-3DHP}, BEHAVE \cref{fig:BEHAVE}, and RICH \cref{fig:RICH}.
All results are estimated by HeatFormer trained on the Human3.6M and MPI-INF-3DHP datasets. The results clearly show that HeatFormer is an occlusion-robust, view-flexible, and generaliziable neural optimizer for multiview HMR.

\section{Calibration}
HeatFormer uses camera extrinsics only for AdaFuse \cite{zhang2020adafuse} and heatmap projection.
HeatFormer can estimate the SMPL parameters without camera extrinsics by skipping AdaFuse and estimating weak-perspective camera parameters instead of the translation calculated with extrinsics.
\cref{tab:calib} shows accuracy comparison between HeatFormer applied to calibrated and uncalibrated cameras for 4 views of 3-iter model.
As the results show, leveraging calibration information leads to higher accuracy, but even uncalibrated HeatFormer achieves reasonable accuracy.

\begin{table}[]
    \centering
    {
    \begin{tabular}[t]{@{}l|ccc@{}}
         \toprule
         & \textbf{MPJPE$\downarrow$} & \textbf{PA-MPJPE$\downarrow$}\\
         \midrule
         calibrated(iter3) & 30.3 & 23.2 \\
         uncalibrated(iter3) & 42.5 & 25.8 \\
         \bottomrule
    \end{tabular}
    }
    \caption{The comparison between calibrated and uncalibrated of 3-iter model.}
    \label{tab:calib}
\end{table}

\section{Ablation Study}
\label{sec:abl}





HeatFormer achieves HMR with neural optimization. We use three forward inferences through the decoder (three unrolled iterations) by default. \cref{tab:ab-iter} shows the accuracy for models trained for different numbers of iterations (1 to 5). The results clearly show that the more iterations the better but with diminishing returns. We empirically found three or four iterations suffice for the HMR accuracy to converge.

\begin{table}
  \centering
  \scalebox{0.9}{
  \begin{tabular}{@{}c|cccc@{}}
    \toprule
    \textbf{the \# of iterations} & \textbf{MPJPE $\downarrow$} & \textbf{PA-MPJPE $\downarrow$} & \textbf{MPVPE $\downarrow$} \\
    \hline
    1 & 34.1 & 27.2 & 43.9 \\
    2 & 30.7 & 25.2 & 39.0 \\
    3 & \cellcolor{yellow!25}28.6 & \cellcolor{yellow!25}23.1 & \cellcolor{yellow!25}36.2 \\
    4 & \cellcolor{red!25}27.2 & \cellcolor{red!25}22.1 & \cellcolor{orange!25}35.2 \\
    5 & \cellcolor{orange!25}27.5 & \cellcolor{orange!25}22.4 & \cellcolor{red!25}34.9 \\
    \bottomrule
  \end{tabular}}
  \caption{Ablation study on the number of forward inferences through the decoder (\ie, number of neural optimization iterations). We train on Human3.6M and MPI-INF-3DHP and test on Human3.6M.}
  \label{tab:ab-iter}
\end{table}

HeatEncoder consolidates the joint heatmaps for each view to extract a rich integrated feature map reflecting the shape and pose of the person observed in the view. This consolidation is essential for the decoder to align the heatmaps through cross-attention and its iterative application to arrive at accurate view-dependent SMPL estimates. The HeatEncoder applied both to the input heatmaps as well as the heatmaps computed from the current SMPL estimate plays a crucial role in making full use of the spatial coordination of joints in each view. We evaluate the effectiveness of HeatEncoder by replacing it with a simple alternative of just taking the max values of the heatmaps across all joints (\ie, per-pixel max-pooling) and applying a heatmap-pre-trained ViT to extract features. 
We train on Human3.6M and MPI-INF-3DHP and test on Human3.6M.
The results shown in \cref{tab:heatmap_ablation} clearly demonstrate the effectiveness of HeatEncoder. 

\begin{table}
    \centering
    \scalebox{0.9}{
    \begin{tabular}{l|cccc}
        \toprule
         & \textbf{MPJPE $\downarrow$} & \textbf{PA-MPJPE $\downarrow$} & \textbf{MPVPE $\downarrow$}\\
        \hline
        w/HeatEncoder & \cellcolor{red!25}28.6 & \cellcolor{red!25}23.1 & \cellcolor{red!25}36.2 \\
        w/o & 30.5 & 25.6 & 38.3 \\
        \bottomrule
    \end{tabular}}
    \caption{Evaluation of the effectiveness of HeatEncoder by replacing it with simple max-pooling and ViT feature extraction (``w/o''). The accuracy is evaluated on Human3.6M. The results clearly show that HeatEncoder is essential for heatmap consolidation in each view.}
    \label{tab:heatmap_ablation}
\end{table}

HeatFormer leverages multiview images and updates SMPL parameters with cross-attention.
As an abalation study of the decoder of HeatFormer, we replace it with simple pooling and MLP.
\cref{tab:ab-heatformer} shows the results which clearly demonstrate the effectiveness of the architecture of HeatFormer.

\begin{table}[]
    \centering
    {
    \begin{tabular}[t]{@{}l|ccc@{}}
        \toprule
        & \textbf{MPJPE$\downarrow$} & \textbf{PA-MPJPE$\downarrow$} \\
        \midrule
        w/HeatFormer & \cellcolor{red!25}28.6 & \cellcolor{red!25}23.1 \\
        w/o & 60.4 & 46.4 \\
        \bottomrule
    \end{tabular}
    }
    \caption{Evaluation of the effectiveness of HeatFormer by replacing it with simple pooling and MLP. The accuracy is calculated on Human3.6M. The results clearly show that the decoder of HeatFormer is essential for neural optimization.}
    \label{tab:ab-heatformer}
\end{table}

Heatmaps computed from the current SMPL model are combined with image features computed from the input views to form decoder queries for cross-attention with the encoder output tokens.
Without these image features, the cross-attention is unlikely to produce meaningful transformations to the heatmaps as the decoder would not know view-correspondences. 
To confirm this, we evaluate the model without combining image features with the view-dependent heatmaps as decoder queries.
Please note that we use image features only for global orientation estimation.
We train on Human3.6M and MPI-INF-3DHP, and test on Human3.6M.
\cref{tab:ab-image} show the results which clearly show that the use of image features with the heatmaps is essential for accurate decoder inference. 

\begin{table}
    \centering
    \scalebox{0.9}{
    \begin{tabular}{l|cccc}
        \toprule
         & \textbf{MPJPE $\downarrow$} & \textbf{PA-MPJPE $\downarrow$} & \textbf{MPVPE $\downarrow$} \\
        \hline
        w/Image Feature & \cellcolor{red!25}28.6 & \cellcolor{red!25}23.1 & \cellcolor{red!25}36.2 \\
        w/o & 42.8 & 33.8 & 56.1 \\
        \bottomrule
    \end{tabular}}
    \caption{Evaluation of the effectiveness of using image features in combination with heatmaps for the decoder queries. The accuracy is evaluated on Human3.6M. Combining the image features is essential for accurate estimation through decoder inference.}
    \label{tab:ab-image}
\end{table}

A key contribution of HeatFormer lies in the adoption of heatmaps as the fundamental representation of pose and their seamless integration in the neural optimization pipeline, which is essential to obtain dense spatial gradients to let it learn to optimize.
To confirm the effectiveness of heatmap representation, we compare with just tokenizing keypoint locations (\ie, directly input and estimate joint coordinates as tokens).
\cref{tab:token-joint} shows that the heatmaps are essential for our high accuracy.

\begin{table}[]
    \centering
    {
    \begin{tabular}[t]{@{}l|ccc@{}}
         \toprule
         & \textbf{MPJPE$\downarrow$} & \textbf{PA-MPJPE$\downarrow$}\\
         \midrule
         Heatmap & \cellcolor{red!25}30.3 & \cellcolor{red!25}23.2 \\
         Joint & 51.2 & 37.0 \\
         \bottomrule
    \end{tabular}
    }
    \caption{Comparison using heatmap representation with direct tokenization of keypoint locations.}
    \label{tab:token-joint}
\end{table}

\begin{figure}
\centering
    \includegraphics[width=\linewidth]{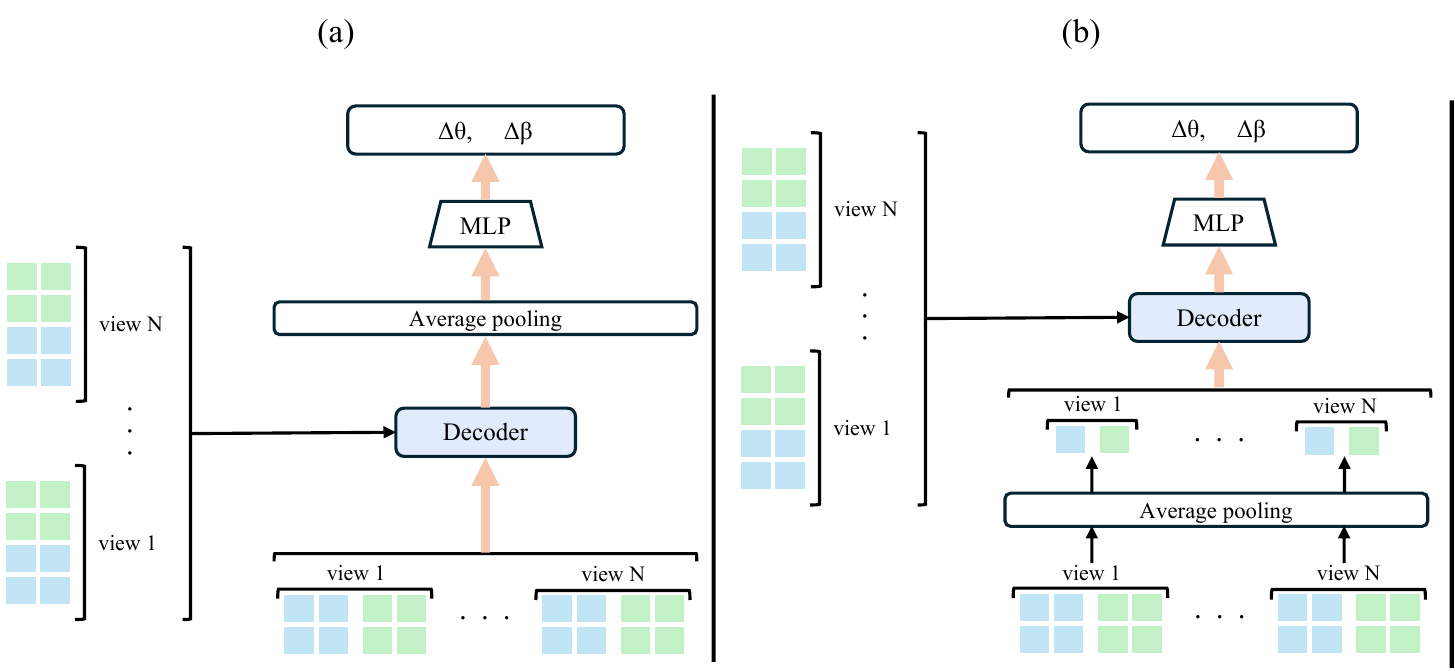}
    \caption{(a) HeatFormer decoder in which the spatial configuration of heatmaps are retained and view-dependent outputs are averaged pooled to compute the SMPL parameters. (b) A variant in which the patchified heatmaps for each view are average pooled before input to the decoder as queries.}
    \vspace{-1\baselineskip}
    \label{fig:ab-ap}
\end{figure}

SMPL parameters are computed from the decoder output after average pooling. Retaining the patchified spatial structure of the decoder queries throughout the cross-attention and only average pooling after it is essential to fully leverage the spatial configuration of the heatmaps across the different views (\cref{fig:ab-ap}(a)). We confirm the importance of retaining this spatial configuration by comparing it with a variant of the decoder where the patchified heatmaps for each view are average pooled before input to the decoder (\cref{fig:ab-ap}(b)) and thus the spatial configuration is dampened. 
We train on the Human3.6M and MPI-INF-3DHP datasets and test on the Human3.6M dataset.
As \cref{tab:ab-ap} shows retaining the spatial configuration of the heatmaps through cross-attention and then consolidating the views via average pooling (\ie, the decoder of HeatFormer) achieves higher accuracy. 

\begin{table}
    \centering
    \scalebox{0.9}{
    \begin{tabular}{l|cccc}
        \toprule
         & \textbf{MPJPE $\downarrow$} & \textbf{PA-MPJPE $\downarrow$} & \textbf{MPVPE $\downarrow$}\\
        \hline
        (a) & \cellcolor{red!25}28.6 & \cellcolor{red!25}23.1 & \cellcolor{red!25}36.2 \\
        (b) & 36.3 & 26.6 & 46.0 \\
        \bottomrule
    \end{tabular}}
    \caption{Comparison of the decoders in \cref{fig:ab-ap}. The HeatFormer decoder which retains the spatial configuration of query heatmaps (a) achieves higher accuracy than the variant that consolidates spatial information of the heatmaps through average pooling before they are input to the decoder as queries (b).}
    \label{tab:ab-ap}
\end{table}

\begin{table}
    \centering
    \scalebox{0.9}{
    \begin{tabular}{l|cccc}
        \toprule
         & \textbf{MPJPE $\downarrow$} & \textbf{PA-MPJPE $\downarrow$} & \textbf{MPVPE $\downarrow$}\\
        \hline
        (a) & \cellcolor{red!25}28.6 & 23.1 & \cellcolor{red!25}36.2 \\
        (b) & 36.3 & \cellcolor{red!25}22.8 & 47.6 \\
        \bottomrule
    \end{tabular}}
    \caption{Comparison between view-dependent estimation (a) and average global estimation (b). Even with the crude averaging, if necessary, HeatFormer's view-dependent estimates can be consolidated without too much loss in accuracy.}
    \vspace{-1.15\baselineskip}
    \label{tab:ab-average}
\end{table}

HeatFormer outputs view-dependent SMPL estimates, \ie, the SMPL parameters explain each image independently. If a single estimate is necessary, we could combine these view-dependent estimates in any way suitable for the downstream task. For instance, if a SMPL model is necessary for a view in between the input views, SMPL estimates of closest views can be combined. A simple approach to consolidating all views would be to average pool them. We can take the average of the pose parameters without the global orientation and also all the shape parameters. \cref{tab:ab-average} shows the results of comparing the accuracies of this average global SMPL estimate and our view-dependent estimates. The view-dependent estimates are naturally more accurate, but even a crude averaging will not loose too much accuracy. In general, for downstream tasks, we can select closest views to achieve higher accuracy than averaging.

%

{
    \small
    \bibliographystyle{ieeenat_fullname}
    \bibliography{main}
}

\end{document}